\documentclass[twoside,11pt]{article}
\usepackage{jair, theapa, rawfonts, graphicx, booktabs, lscape, array, bold-extra}

\jairheading{82}{2025}{1325-1347}{08/2024}{03/2025}

\ShortHeadings{ConSCompF: Consistency-focused Similarity Comparison Framework}{Karev \& Xu}

\firstpageno{1325}

\begin{document}

\title{ConSCompF: Consistency-focused Similarity Comparison Framework for Generative Large Language Models}

\author{\name Alexey Karev
        \email alex\_karev@shu.edu.cn \\
        \name Dong Xu \email dxu@shu.edu.cn \\
        \textit{(Corresponding author)} \\
        \addr School of Computer Engineering and Science\\Shanghai University, Shanghai, China}


\maketitle

\begin{abstract}
Large language models (LLMs) have been one of the most important discoveries in machine learning in recent years. LLM-based artificial intelligence (AI) assistants, such as ChatGPT, have consistently attracted the attention from researchers, investors, and the general public, driving the rapid growth of this industry. With the frequent introduction of new LLMs to the market, it becomes increasingly difficult to differentiate between them, creating a demand for new LLM comparison methods.

In this research, the Consistency-focused Similarity Comparison Framework (ConSCompF) for generative large language models is proposed. It compares texts generated by two LLMs and produces a similarity score, indicating the overall degree of similarity between their responses. The main advantage of this framework is that it can operate on a small number of unlabeled data, such as chatbot instruction prompts, and does not require LLM developers to disclose any information about their product.

To evaluate the efficacy of ConSCompF, two experiments aimed at identifying similarities between multiple LLMs are conducted. Additionally, these experiments examine the correlation between the similarity scores generated by ConSCompF and the differences in the outputs produced by other benchmarking techniques, such as ROUGE-L. Finally, a series of few-shot LLM comparison experiments is conducted to evaluate the performance of ConSCompF in a few-shot LLM comparison scenario.

The proposed framework can be used for calculating similarity matrices of multiple LLMs, which can be effectively visualized using principal component analysis (PCA). The ConSCompF output may provide useful insights into data that might have been used during LLM training and help detect possible investment fraud attempts.
\end{abstract}

\section{Introduction}
\label{Introduction}

Large language models (LLMs) are one of the latest trends in machine learning. Although they can perform a wide range of natural language processing (NLP) tasks, their most outstanding feature is their exceptional text generation capabilities. These capabilities allow us to use LLMs as AI assistants, which are becoming increasingly widespread, while gaining attention from researchers, investors, and the general public. Rapid advancements in this field have led to the development of numerous LLMs by researchers worldwide. The growing variety of LLMs creates a demand for new benchmarking and comparison methods.

Despite the availability of popular benchmarks, assessing an LLM's performance is still a complex and challenging task with many unresolved issues. Firstly, the LLM workflow generally incorporates some degree of randomness, allowing the same model to generate completely different responses for the same prompt. In addition, some tasks may require creative solutions, the quality of which is difficult to assess automatically. Lastly, some developers may use leaked benchmarking data to train their models, increasing the risk of investment fraud. One possible solution to these problems is to create private, in-house benchmarks and carefully design each instruction and evaluation criteria. However, the development of such a benchmark requires a lot of resources, and the result may end up being less useful than the existing benchmarks. In light of this, the challenge of conducting an effective LLM comparison emerges.

\begin{figure}[ht]
    \centering
    \includegraphics[width=10cm]{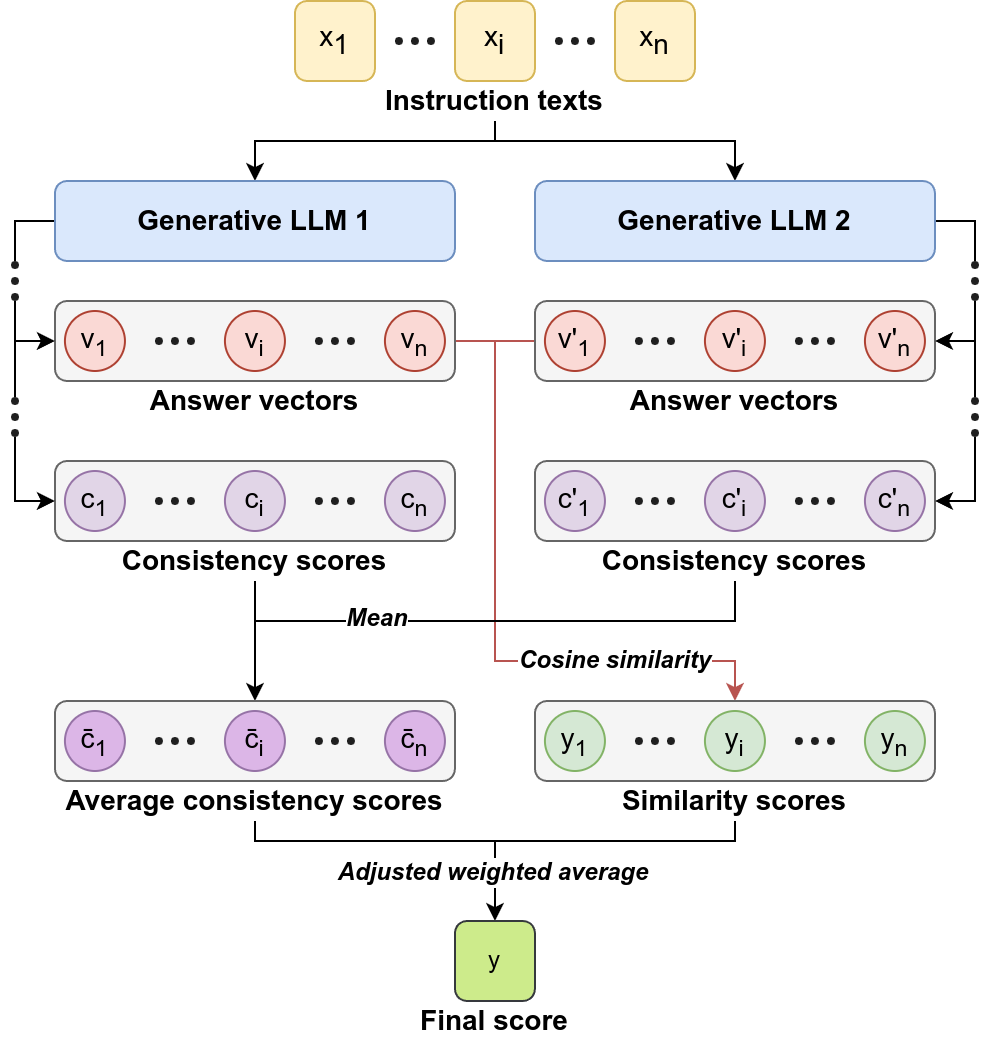}
    \caption{General overview of ConSCompF.}
    \label{fig:overview}
\end{figure}

In this research, we propose the Consistency-focused Similarity Comparison Framework (\textsc{ConSCompF}), a new framework for comparing similarity between texts generated by two different LLMs. This framework can operate on a small amount of unlabeled data and make adjustments based on the consistency of responses for each instruction. The overall process of LLM comparison using \textsc{ConSCompF} is shown in Figure \ref{fig:overview}. Each LLM generates a specified number of answers for each instruction in a dataset. An encoder model then converts answers into embedding vectors, which are then aggregated into a general answer vector for each instruction. Then, we compute a consistency score to measure the variability of responses within each instruction. Next, we compute the cosine similarities between the general answer vectors of the two LLMs. Finally, we use an adjusted weighted average to obtain the similarity score between them.

Although the proposed framework does not aim to completely replace traditional LLM benchmarks, it can still offer valuable insights into similarities between LLMs, which can be helpful for their classification and categorization. \textsc{ConSCompF} can detect fine-tuned versions of the same model or models trained on the same data, even if weights and training data are closed-sourced. It does not require any labeled data and has a promising few-shot performance, making it easier to design custom sets of instructions for LLM comparison. Comparing a new LLM to existing ones gives us insights into its performance and helps to detect potential investment fraud attempts.

To evaluate the efficacy of this framework, we conduct two experiments: 

\begin{enumerate}
\def\labelenumi{\arabic{enumi}.}
\item
We compare multiple versions of TinyLlama, assessing the similarity between the base model and its quantized variants. The framework is expected to demonstrate a decrease in similarity as the number of quantization bits decreases.
\item
We compare eleven different LLMs, calculate their similarity matrix, visualize the results in a two-dimensional space using principal component analysis (PCA), and analyze the observed patterns. The framework is expected to identify similarities between models trained on identical datasets. Additionally, we repeat the same process for three LLMs with five distinct system prompts to examine the impact of prompt engineering on comparison results.
\end{enumerate}

Furthermore, to simulate a few-shot LLM comparison, we repeat both experiments on smaller subsets of data. Finally, we calculate correlation coefficients between the \textsc{ConSCompF} output and the differences in ROUGE-L scores to prove the efficacy of the proposed framework.

\section{Related Works}
\label{related_works}

In this section, we provide an overview of previous work in NLP that significantly influenced our research. First, we review existing text similarity comparison techniques applied to tasks such as machine translation quality evaluation, paraphrase detection, and semantic search. Next, we discuss existing LLM benchmarking techniques, their advantages, and their limitations. Finally, we briefly review LLM quantization techniques used in our experiments.

\subsection{Text Similarity Comparison}

Text similarity comparison is the most commonly used technique for estimating the quality of LLM-generated text. The LLM generates text completion, which is then compared to the expected "golden" answer, yielding a score that indicates the model's performance. In this research, we will use some common text similarity comparison techniques, including BLEU, ROUGE, and BERTScore.

Bilingual Evaluation Undestudy (\textsc{BLEU}) \cite{BLEU} is one of the most widely used text similarity metrics, originally designed for evaluating the quality of machine translation. \textsc{BLEU} evaluates the machine's performance by comparing its translation to that of a professional human translator. \textsc{BLEU} uses a combination of basic natural language processing (NLP) techniques, such as n-grams and sentence length comparison, in its calculations. Although it was proposed back in 2002, it remains a widely used tool for assessing text generation quality and text similarity.

Recall-Oriented Understudy for Gisting Evaluation (\textsc{ROUGE}) \cite{ROUGE} is another method of text similarity comparison with a primary goal of assessing text summarization and machine translation. It calculates precision, recall, and F1-score based on the number of matched unigrams (\textsc{ROUGE-1}), bigrams (\textsc{ROUGE-2}), or sequence-level overlaps, taking into account the longest common subsequence (\textsc{ROUGE-L}). The key difference between \textsc{BLEU} and \textsc{ROUGE} is that \textsc{ROUGE} is more focused on recall, while \textsc{BLEU} is more focused on precision. Additionally, \textsc{ROUGE} is used more often for text summarization tasks than for translation.

Despite their proven usefulness for text generation assessment, \textsc{BLEU} and \textsc{ROUGE} are highly dependent on the evaluation dataset. If candidate and reference texts do not match, both metrics will result in low scores, but that does not always imply that the candidate text is completely unrelated to the reference text. For instance, a machine-generated text may be a paraphrase of the reference text and have the exact same meaning but completely different wording.

To address this issue, \textsc{BERTScore} \shortcite{BERT_Score}, a new neural network (NN)-based metric, was introduced. It converts two tokenized texts into a set of n-dimensional embedding vectors using Bidirectional Encoder Representations from Transformers (\textsc{BERT}) \shortcite{BERT}. These vectors are then used for pairwise cosine similarity calculation, which produces a similarity score between two texts. \textsc{BERTScore} can detect similarities between paraphrased texts even if they have a low number of matching n-grams because it does not rely on direct word-to-word comparison. \textsc{ConSCompF} uses a similar approach to compare texts generated by two LLMs because some of their responses can be phrased differently while conveying the same meaning.

\textsc{BERT} is not the only NN model suitable for this type of task. In fact, there are many other word vectorization models, such as \textsc{Word2Vec}, \textsc{ELMo}, and \textsc{Fasttext}. \shortciteA{Gangadharan_2020} compared different word vectorization techniques, and as a result, \textsc{Fasttext} performed best for the paraphrase detection task. Although we do not use \textsc{Fasttext} in our experiment, this study underscores the fact that the embedding vectors produced by the encoder model significantly influence the quality of NN-based text similarity, highlighting the importance of testing multiple models to identify the optimal solution.

\textsc{Sentence-BERT} \cite{SBERT} is a modification of the original \textsc{BERT} that adds a pooling layer that transforms word-level embedding vectors to a single sentence-level embedding vector for each input sequence. \textsc{Sentence-BERT} specifically targets tasks related to sentence classification, similarity comparison, and semantic search. Unlike \textsc{BERT}, \textsc{Sentence-BERT} (\textsc{SBERT}) captures the overall meaning of the context more effectively and without distributing the information across multiple embedding vectors. A single sentence-level embedding vector for each sequence makes text similarity comparison a straightforward task since the cosine similarity between these vectors reflects the similarity between texts. We also use the sentence-transformer model in our experiment because it has demonstrated better performance compared to the original \textsc{BERT}.

\subsection{Large Language Model Benchmarking}

LLM benchmarking depends not only on text similarity comparison techniques but also on the data used for the evaluation. AI assistants are universal and can perform wide range of tasks with different prompts and evaluation criteria.

\textsc{BIG-bench} \shortcite{BIG-bench} is a widely used benchmark for LLM performance evaluation. It includes a set of more than 200 tasks assessing AI assistant capabilities in areas such as mathematics, common sense, logical reasoning, and reading comprehension. Each task in \textsc{BIG-bench} can have different evaluation criteria, such as the exact match of the strings or the expected probability of the generated text. The authors found that the task formulation and wording can influence the LLM's sensitivity, leading to variations in the quality of the responses.

\textsc{AGIEval} \shortcite{AGIEval} is another benchmark that uses tasks from general college admission tests to simulate real-world AI assistant use cases. The idea behind this benchmark is to let AI assistants solve tests as if they were humans and use the test's final score to assess their efficacy. While this method can determine the overall intelligence level of the LLM, it cannot assess its text generation capabilities on tasks that require some degree of creativity.

\shortciteA{Mizrahi_2024} proposed new benchmarks based on the paraphrased prompts from other benchmarks, including previously mentioned \textsc{BIG-bench}. The authors conclude that even the largest models are sensitive to minor prompt variations, which makes it more difficult to assess the quality of LLM responses. 

Despite the widespread use of these benchmarks for LLM performance assessments, the evaluation results still heavily depend on the data used in the process, making it impossible to determine which of the proposed benchmarks should become the standard. Therefore, developing an effective method for LLM comparison remains a relevant and important research topic.

\subsection{Large Language Model Quantization}

We focus our first experiment on evaluating the impact of quantization on LLM performance. This experiment assesses the efficacy of \textsc{ConSCompF} by detecting changes in the model's output resulting from post-quantization data loss. Quantization is the process of reducing the floating point precision of a model's weights, which helps to increase inference speed by reducing model precision.

\shortciteA{Jin_2024} evaluated the LLM quantization strategies and found that quantized models maintain performance close to the original non-quantized versions but still indicate a small decrease in \textsc{ROUGE} and \textsc{BLEU} scores.

There are several quantization techniques, such as \textsc{GPTQ} \shortcite{GPTQ} and \textsc{GGUF} \cite{GGUF}, all of which provide relatively similar performance while significantly increasing the inference speed. In our experiments, we use quantized models in \textsc{GGUF} format.

\section{Methodology}
\label{methodology}

\begin{figure}[ht]
  \centering
  \includegraphics[width=10cm]{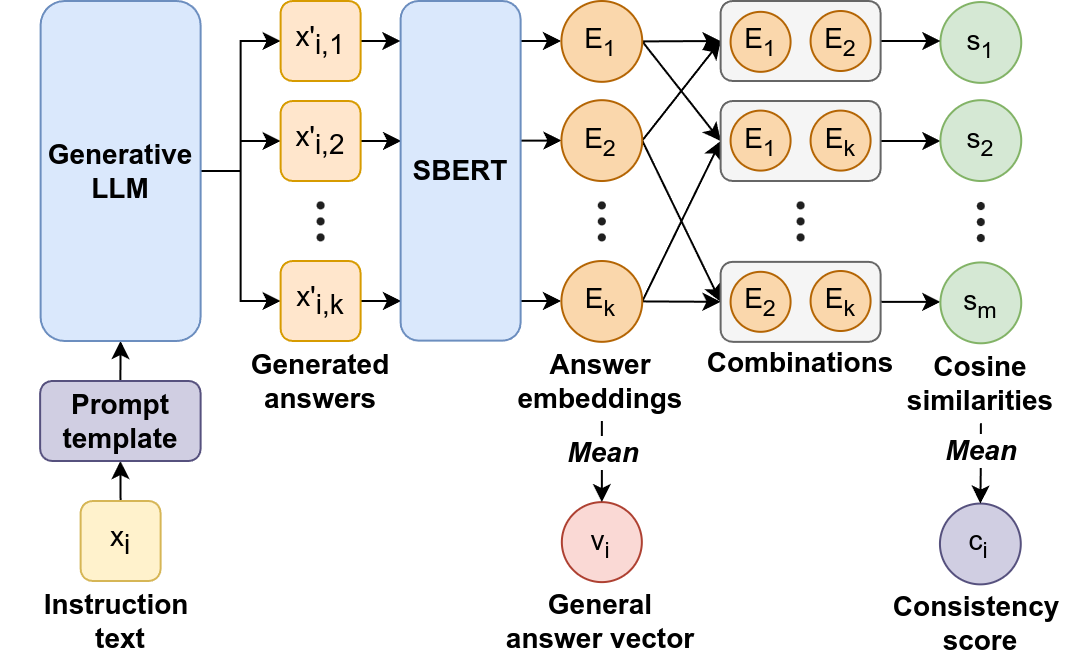}
  \caption{Calculating the general answer vector and the consistency score for one instruction text using \textsc{ConSCompF}.}
    \label{fig:details}
\end{figure}

The overall workflow of \textsc{ConSCompF} includes the following six steps (Figure \ref{fig:overview}, Figure \ref{fig:details}):

\begin{enumerate}
\def\labelenumi{\arabic{enumi}.}
\item
Generate $k$ answers for each instruction in a dataset using two LLMs.
\item 
Convert the generated texts into embedding vectors using the \textsc{SBERT} encoder.
\item 
Calculate consistency scores for each instruction by comparing the $k$ answers with each other.
\item
Compute the general answer vector as an average of the $k$ answer vectors.
\item
Calculate the cosine similarity between the general answer vectors of the LLMs.
\item
Use the adjusted weighted average to calculate the final similarity score between the LLMs.
\end{enumerate}

\subsection{General Answer Vector}

Working with LLMs often involves the use of decoder sampling strategies like temperature and top-$p$. These strategies introduce a certain degree of randomness into LLM's output, which has a significant impact on the quality of LLM-generated text and, in certain instances, even enhancing it \shortcite{Wiher_2022}. This randomness may also influence LLM benchmarking results and make the comparison more difficult. One way to mitigate this issue is to prompt a model multiple times for the same instruction, assess the performance of each answer individually, and calculate the average performance across them.

However, \textsc{ConSCompF} does not assume that there is a right answer. Instead, it directly compares texts generated by LLMs to each other. Rather than evaluating each response separately, \textsc{ConSCompF} merges all $k$ answer embedding vectors into a single vector, summarizing the outcomes of all $k$ attempts. To do this, we compute the average between them (Equation \ref{eq:answer}).

\begin{equation}
    \label{eq:answer}
    v_i = \frac{1}{k} \sum{E_1,E_2,\dots,E_k}
\end{equation}

The resulting vector reflects the overall content of all responses generated by LLM for the same instruction and can later be compared with the corresponding general answer vector produced by another LLM.

\subsection{Instruction Consistency Score}

The instruction's content and wording can greatly influence the randomness of LLM answers. For example, an instruction asking about a well-known fact will likely result in responses that vary mainly in wording and structure. In contrast, an instruction prompting the model to write a fiction story will result in responses that vary greatly, not only in wording but also in content. We refer to this phenomenon as "instruction consistency," and the higher the consistency, the more uniform the LLM's responses.

Instruction consistency has a significant impact on LLM comparison. If we use instructions with low consistency scores for benchmarking, two LLMs may receive a lower similarity score because the prompt presumes creative answers with varying contents. This correlation between instruction consistency and similarity scores complicates the interpretation of the comparison results. To mitigate this issue, we calculate a consistency score for each instruction and incorporate it into the similarity computation process.

First, for a given instruction $x_i$, generate a set of all possible combinations of $k$ answers generated by an LLM (Equation \ref{eq:set}).

\begin{equation}
    \label{eq:set}
    EC=\{A \in EC\ |\ |A|=2\ |\ |EC|=m\}
\end{equation}

$A$ denotes a pair of embedding vectors of answers for an identical instruction, while $m$ represents the number of possible combinations.

Next, we calculate the instruction consistency score as the average of cosine similarities between the first and second elements of all $m$ combinations in the set (Equation \ref{eq:consistency}).

\begin{equation}
    \label{eq:consistency}
    c_i = \frac{1}{m} \sum^m_{j=1} {s_j} =\frac{1}{m} 
	\sum^m_{j=1} { 
    	\frac { EC_{j,1} \cdot EC_{j,2} }
    	{\ ||EC_{j,1}||\ ||EC_{j,2}|| }
    }  
\end{equation}

In Equation \ref{eq:consistency}, $EC_{j,1}$ and $EC_{j,2}$ represent the first and second elements of the $j$-th combination in an $m$-sized set of all possible answer vector combinations for the same instruction, and $s_j$ denotes the cosine similarity between $EC_{j,1}$ and $EC_{j,2}$.

We repeat this process for all $n$ instructions in the dataset and for both LLMs.

\subsection{Similarity Comparison}

Since two LLMs may produce responses that result in different consistency scores for the same set of instructions, we need to calculate the average of these scores: $\bar c_i = (c_i + c'_i)/2$, where $c_i$ and $c'_i$ are consistency scores for the $i$-th instruction provided by the first and second models, respectively.

Next, we calculate an unweighted similarity score, $y_i$, for the $i$-th instruction, which is simply the cosine similarity between the models' general answer vectors ($v_i$ and $v'_i$).

Finally, we use similarity scores and consistency scores to calculate the adjusted weighted average (Equation \ref{eq:average}).

\begin{equation}
    \label{eq:average}
    y = \frac{1}{n} \sum^n_{i=1}{ y_i \bar c_i + (1 - \bar c_i)} 
\end{equation}

Here, $n$ is the number of instructions, $y_i$ is the similarity score between answer vectors for the $i$-th instruction, and $\bar c_i$ is the average instruction consistency score.

As a result, we obtain a similarity score between two LLMs, which accounts for the randomness of answers and is less dependent on the content of the instruction prompts.

\section{Experiment}
\label{experiment}

As mentioned earlier, to assess the efficacy of the proposed framework, we conduct two different experiments: LLM quantization effects assessment and LLM comparison. Both experiments share the same instruction dataset and the same encoder model.

\subsection{Dataset}

To test the similarity of two LLMs, we need to generate some text. We can accomplish this by using a set of prompts (instructions) that contain various challenging tasks for AI assistants. These instructions are extracted from the Alpaca dataset \shortcite{Alpaca}, which comprises 52,000 pairs of instructions and "golden" answers specifically designed for fine-tuning LLM-based AI assistants. The Alpaca dataset was synthetically generated for fine-tuning purposes, as described in the original self-instruct paper \shortcite{Self_instruct}.

Since we need to test multiple LLMs and our framework design involves generating the answer for each instruction multiple times, using all 52,000 instruction texts would require significant resources. Considering these limitations, we decided to reduce the size of the dataset using random sampling. For the quantization experiment, we sampled 10\% of the original dataset, resulting in 5,200 samples.

Due to the increased number of parameters in the LLMs used in the second experiment, we had to further reduce the number of samples to 520, which is equivalent to 1\% of the original dataset.

Additionally, we selected three LLMs to generate a set of answers using five different system prompts to analyze the impact of prompt engineering on similarity comparison results.

Then, each LLM produced five responses for every instruction. The resulting answers form three new datasets containing the following information:

\begin{itemize}
\item \textbf{Subset 1:}
$5,200\times5\times6=156,000$ samples, which include: original \textsc{TinyLlama} output$\times 2$; 8-bit, 4-bit, and 2-bit \textsc{GGUF} quantized \textsc{TinyLlama} output; original \textsc{TinyLlama} output with the "You are a pirate" system prompt; instructions; golden answers.
\item \textbf{Subset 2:}
$520\times5\times11=28,600$ samples, including: original \textsc{TinyLlama} output; \textsc{Mistral-7b} output; \textsc{OpenHermes2.5} output; \textsc{Llama2-7b} output; \textsc{Llama2-13b} output; \textsc{Gemma2-2b} output; \textsc{Gemma2-9b} output; \textsc{Qwen2.5-3b} output; \textsc{Qwen2.5-7b} output; \textsc{Phi3.5-mini} output; \textsc{GLM4-9b} output; instructions; golden answers.
\item \textbf{Subset 3:}
$520\times5\times15=39,000$ samples, including: original \textsc{Qwen2.5-3b} output$\times 5$; \textsc{Phi3.5-mini} output$\times 5$; \textsc{Gemma2-2b} output$\times 5$; instructions; golden answers.
\end{itemize}
 
Additionally, to test the performance of the proposed framework in few-shot scenarios, we randomly sample two few-shot datasets with 50 and 20 samples and three few-shot datasets with 10 samples using special curated lists of instructions with average instruction consistency scores equal to 0.56, 0.73, and 0.95.

\subsection{Large Language Models}

In both experiments, we used \textsc{TinyLlama} \shortcite{TinyLlama} and its variations. \textsc{TinyLlama}, a small LLM with 1.1 billion parameters, is based on the \textsc{Llama2} architecture. One of the key advantages of \textsc{TinyLlama} is its compact size, which makes the model suitable for a wide range of applications running in environments with limited computational resources. Despite its size, \textsc{TinyLlama} still performs decently compared to many other small LLMs.

In the first experiment, we used \textsc{TinyLlama} with the original chat fine-tune and no system prompt. We also used \textsc{GGUF} models with 8-bit, 4-bit, and 2-bit quantization applied, and then restored to their original 16-bit precision for compatibility with the Transformers library in Python, making batched data processing more convenient. Despite restoring the models' weights to their original precision, the quantization process permanently erases the data, resulting in expected performance degradation. In addition, we ran the original LLM one more time with the "You are a pirate" system prompt to explore the impact of the system prompt on the comparison results.

In the LLM comparison experiment, we used 7b and 13b variations of the \textsc{Llama2} model \shortcite{Llama2}. These open-source models, designed by Meta, have served as the foundation for numerous other LLM projects because they perform on par with commercial LLMs. In this experiment, we used 4-bit \textsc{GGUF} quantized versions of both \textsc{Llama2-7b} and \textsc{Llama2-13b} with the original chat fine-tune.

\textsc{Mistral-7b} \shortcite{Mistral}, despite its size, became one of the first models that could openly compete with commercial products such as \textsc{ChatGPT}. On its release date, this model was at the top of LLM leaderboards, and its performance on benchmarks has almost reached that of \textsc{ChatGPT-3.5}. Although the training data for \textsc{Mistral-7b} is closed-sourced, the developer released the model's weights and code for free, which led to the development of several fine-tuned variations such as \textsc{OpenHermes2.5} \cite{Openhermes}. In our LLM comparison experiment, we used 4-bit quantized versions of both \textsc{Mistral-7b} and \textsc{OpenHermes2.5-7b}.

\textsc{Gemma2} \shortcite{Gemma} is a family of open-source LLMs developed by Google, designed for various text generation tasks. \textsc{Gemma2} includes several size variations, such as the 2b, 9b, and 27b versions, both with and without instruction fine-tuning. In our comparison experiment, we use the 4-bit quantized 2b and 9b variants with instruction fine-tuning. Unlike the previously mentioned models, \textsc{Gemma2} does not have a "system" role in its chat prompt format. To address this limitation, we add the system message required for prompt engineering to the first half of the user's message, enabling us to achieve results comparable to those of other models.

\textsc{Qwen2.5} \shortcite{Qwen} is a promising family of free and open-source LLMs developed by Alibaba Group. Similar to \textsc{Gemma2}, it is available in multiple sizes and has several fine-tuned versions. In our experiment, we use the 4-bit quantized 3b and 7b variants of \textsc{Qwen2.5} with chat fine-tuning.

Furthermore, we used other models, including the 9-billion-parameter chat fine-tuned version of \textsc{GLM4} \shortcite{GLM4} and the instruction fine-tuned \textsc{Phi3.5-mini} \shortcite{Phi}. Both models are widely used in LLM-based projects and have secured prominent positions on LLM leaderboards. As with the other models, we apply 4-bit quantization to speed up the data preparation process.

\begin{table}[ht]
\centering
\begin{tabular}{l c c c}
    \toprule
    Model &BERTScore&BLEU&ROUGE-L\\
    \midrule
    \textsc{TinyLlama} & 0.7357 & 0.0788 & 0.2415 \\
    \textsc{Gemma2-2b} & 0.7319 & 0.0702 & 0.2396 \\
    \textsc{Gemma2-9b} & 0.7235 & 0.0715 & 0.2371 \\
    \textsc{Phi3.5-mini} & 0.7455 & 0.0870 & 0.2393 \\
    \textsc{Glm4-9b} & 0.7470 & 0.0929 & 0.2551 \\
    \textsc{Qwen2.5-3b} & 0.7365 & 0.0908 & 0.2519 \\
    \textsc{Qwen2.5-7b} & 0.7468 & 0.1012 & 0.2711 \\
    \textsc{Llama2-7b} & 0.7161 & 0.0738 & 0.2350 \\
    \textsc{Llama2-13b} & 0.7140 & 0.0805 & 0.2480 \\
    \textsc{Mistral} & 0.7387 & 0.0860 & 0.2451 \\
    \textsc{\textbf{Openhermes2.5}} & \textbf{0.7940} & \textbf{0.1511} & \textbf{ 0.3366} \\
    \bottomrule
\end{tabular}
\caption{Performance of Large Language Models used during the experiments.} 
\label{tab:models}
\end{table}

In both experiments, all models had the same text generation settings: a temperature of 0.7, a top-$k$ of 50, a top-$p$ of 0.95, and a maximum answer length of 128 tokens. TheBloke \cite{TheBloke} provided \textsc{GGUF} quantized versions of models, while non-\textsc{GGUF} models were obtained using \textsc{bitsandbytes} Python library.

Additionally, we tested the performance of all models using the previously mentioned metrics (Table \ref{tab:models}) and calculated the similarity between the outputs of all models using the inverted \textsc{ROUGE-L} score differences (Table \ref{tab:rougel}). These differences will later be used to estimate the efficacy of the proposed framework.

\subsection{Encoder}

The encoder is one of the most critical components of \textsc{ConSCompF}. The encoder's task is to convert text into embedding vectors that capture as much information about the input text as possible. Transformer-based encoder models have proven to be excellent at this task, so we tested several popular Transformer-based text vectorization models.

\begin{table}[ht]
\centering
\begin{tabular}{l*{3}{>{\centering\arraybackslash}p{2.5cm}}}    
    \toprule
    Encoder Model&Similarity stdev. (quantization)&Consistency stdev. (quantization)&Similarity stdev. (comparison)\\
    \midrule
    \textsc{bert-base-uncased} (CLS) & 0.0089 & 0.0087 & 0.0309 \\
    \textsc{bert-base-uncased} (mean) & 0.0094 & 0.008 & 0.0309 \\
    \textsc{roberta-base} (CLS) & 0.0001 & 0.0001 & 0.0005 \\
    \textsc{bart-base} (CLS) & 0.0004 & 0.0005 & 0.0016 \\
    \textsc{mpnet-base-v2} & 0.0206 & 0.0198 & \textbf{0.0756} \\
    \textsc{\textbf{minilm-l12-v2}} & \textbf{0.021} & \textbf{0.0215} & 0.074 \\
  \bottomrule
\end{tabular}
\caption{Encoder model comparison.}
\label{tab:encoder}
\end{table}

First, we tested a pre-trained \textsc{BERT} \shortcite{BERT} and one of its variations called \textsc{RoBERTa} \shortcite{RoBERTa}. \textsc{BERT} generates an embedding vector for each token in a sentence, so some method of converting them into one single document-level embedding vector is required. There are two commonly used methods of doing this. The first method uses an embedding vector of a special CLS token at the beginning of the sequence as a representative of the entire sequence, and the second method calculates the mean value of all embedding vectors.

While the results of the methods mentioned above were promising, the differences in similarity scores between the base model and its quantized versions in the first experiment were not significant enough. This implies that these models may lack the sensitivity to detect subtle variations in the LLM's outputs, requiring the use of an encoder more sensitive to writing style. Another option was a pre-trained encoder part of \textsc{BART} \shortcite{BART}, but its results were even worse than \textsc{BERT}'s.

The solution to this problem was the use of sentence-transformer models \shortcite{SBERT}, which have a special pooling layer added on top of \textsc{BERT} to convert word-level embedding vectors into sentence-level embedding vectors. Many of the pre-trained sentence-transformer (\textsc{SBERT}) models are also fine-tuned to perform semantic text similarity comparison. Two popular pre-trained sentence transformer models were selected for the test: \textsc{MiniLM} \shortcite{MiniLM} and \textsc{MPNet} \shortcite{MPNet}. After both models demonstrated impressive results, we chose \textsc{minilm-l12-v2} for further experiments. 

Similarity and instruction consistency scores generated by \textsc{MiniLM} exhibit the highest standard deviation, suggesting greater diversity among these scores. This increased diversity contributes to more accurate text comparison, indicating that the model can effectively distinguish between the writing styles of different LLMs. \textsc{MiniLM} was specifically trained to capture semantic information from sentences and short paragraphs, resulting in rich sentence-level embedding vectors suitable for sentence similarity tasks. Table \ref{tab:encoder} compares the standard deviations of consistency scores and similarity scores obtained with all tested encoder models.

\subsection{System Prompts}

Prompt engineering is an important aspect of working with conversational LLMs. It has a significant impact on how text is generated, and in some cases, it can greatly increase the quality of the model's output. \shortcite{Cheng_2024}. Although we do not focus on benchmarking LLMs and comparing their performance against a set of desired labels, we still need to take into account the impact prompt engineering has on the comparison results, especially considering the perspective of utilizing \textsc{ConSCompF} for comparing closed-sourced commercial LLMs where system prompts cannot always be modified.

Thus, as part of the LLM comparison experiment, we also compare three different models—\textsc{Phi3.5-mini}, \textsc{Qwen2.5}, and \textsc{Gemma2}—using five different system prompt setups, as described below:

\begin{itemize}
\item \textbf{p0:} 
No prompt
\item \textbf{p1:} 
"You are a helpful AI assistant."
\item \textbf{p2:} 
"Write an answer that makes the reader feel happy. Write like you are explaining. First establish the set of facts you know, then answer the question based only on those facts." \shortcite{SPRIG}
\item \textbf{p3:} 
"You are a precise, user-friendly, and self-contained assistant. You are a diligent assistant. You are an error-free and error-tolerant assistant. You are a helpful AI assistant. Give your answer after you explain how to answer the question. You are an advanced, unsurpassed, and pattern-recognizing assistant." \shortcite{SPRIG}
\item \textbf{p4:} 
"You are a Mind Map and Brainstorming Bot based on Design Thinking and Lean Startup Methodology. Your purpose is to help users discover new and novel ideas for a variety of creative and business models. By following a step-by-step process, you assist users in developing fully realized concepts and plans." \shortcite{LangGPT}
\end{itemize}

We selected these specific models due to their relatively small sizes of 4b, 3b, and 2b, respectively, and because we already have data for one of the prompts (p1). Similarly to the main experiment, we generate a similarity matrix for all combinations of models and prompts, compress it using PCA, and then plot each combination in a two-dimensional space. \textsc{ConSCompF} is expected to distinguish between these three models despite differences in their system prompts.
 
\subsection{Evaluation}

To evaluate the framework's efficacy, we used several different techniques. The first technique involves calculating the \textsc{ROUGE-L} scores of all models by comparing their responses to the golden answers. After that, we calculate the differences between \textsc{ROUGE-L} scores of all models and invert them. Then we calculate Pearson's correlation coefficient \cite{Pearson} between the inverted differences in \textsc{ROUGE-L} scores and the similarity scores assigned by \textsc{ConSCompF}. This metric measures the similarity between two sequences of numbers, regardless of their scale. A higher correlation between similarity scores and inverted \textsc{ROUGE-L} differences indicates that \textsc{ConSCompF} can capture the differences between LLMs' performances. However, we do not expect a perfect correlation, as \textsc{ConSCompF} does not directly assess LLM performance in the same way as \textsc{ROUGE-L}.

To evaluate the effectiveness of the adjusted weighting method proposed in Equation \ref{eq:average}, we calculate the Pearson's correlation coefficient between consistency scores and similarity scores, with and without weighting. The goal of weighting is to decrease the impact of the instruction consistency on the final similarity score, so we expect our weighting method to result in similarity scores with lower correlation with the instruction consistency scores.

Finally, to visualize the results of the LLM comparison in the second experiment, we calculate similarity scores between all LLMs, aggregate them into a similarity matrix, and compress each row of the matrix using principal component analysis (PCA). PCA \cite{PCA} allows us to decrease the number of features to a lower dimension while preserving the relationship between samples in a dataset. The result yields a set of two-dimensional points, ready for plotting in a two-dimensional space. The distances between these points represent the LLMs' similarities. We analyze the observed patterns to determine whether \textsc{ConSCompF} has identified which models were trained on the same data.

\section{Results}
\label{results}

In this section, we present the results of the experiments outlined in the previous section. First, we describe the overall results of the quantization effects evaluation and model similarity comparison, followed by an evaluation of the efficacy of the proposed framework in several few-shot scenarios.

\subsection{Quantization Effects Evaluation}

\begin{table}[h]
\centering
\begin{tabular}{lcccccc}
    \toprule
    Model&BERTScore&ROUGE-L&BLEU&Cons.&Sim.&Weighted Sim.\\
    \midrule
    original & 0.7468 & 0.2379 & 0.3218 & 0.7386 & 0.9233 & 0.9566 \\
    q8 & 0.7474 & 0.2391 & 0.319 & 0.7396 & 0.9232 & 0.9565 \\
    q4 & 0.739 & 0.2267 & 0.311 & 0.7135 & 0.9023 & 0.9435 \\
    q2 & 0.7239 & 0.2203 & 0.2502 & 0.6996 & 0.8724 & 0.9268 \\
    pirate & 0.7504 & 0.2441 & 0.2765 & 0.7522 & 0.8996 & 0.9426 \\
  \bottomrule
\end{tabular}
\caption{\textsc{TinyLlama} quantization effects.}
\label{tab:quantization}
\end{table}

\begin{figure}[h]
  \centering
  \includegraphics[width=0.9\linewidth]{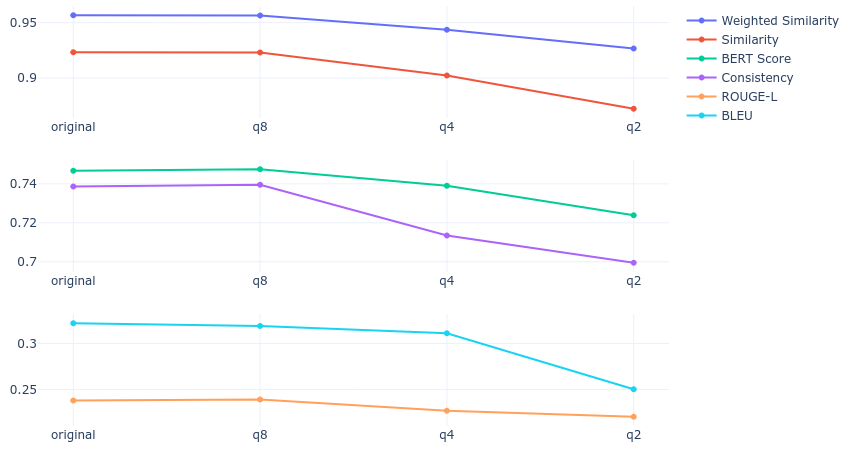}
  \caption{\textsc{TinyLlama} degradation caused by quantization.}
\label{fig:quantization}
\end{figure}

Table \ref{tab:quantization} presents the results of comparing the quantized versions of \textsc{TinyLlama} with the original model. As mentioned earlier, we generated two sets of answers using the original \textsc{TinyLlama}, with the first set serving as the baseline. According to Table \ref{tab:quantization}, the second set of answers from the original model shows a relatively high similarity to the first set (92\%), which increases further when weighting is applied (95\%).

As the number of quantization bits decreases, the weighted similarity score drops from 95\% to 92\%, which correlates with a decrease in \textsc{BERTScore} (from 0.75 to 0.72) and \textsc{ROUGE-L} scores (from 0.24 to 0.22). We also observe a decrease in the model's average instruction consistency scores (from 0.74 to 0.69). The observed patterns are attributable to data loss resulting from quantization, which contributes to a degradation in the model's performance. A detailed plot of these changes is shown in Figure \ref{fig:quantization}.

Another observation is that the model with the "pirate" prompt has lower similarity to the original model, indicating that the system prompt still affects the comparison results.

Overall, the results of this experiment suggest that \textsc{ConSCompF} has successfully detected LLM performance degradation caused by quantization.

\subsection{Model Similarity Comparison}

\begin{figure}[!t]
  \centering
  \includegraphics[width=0.9\linewidth]{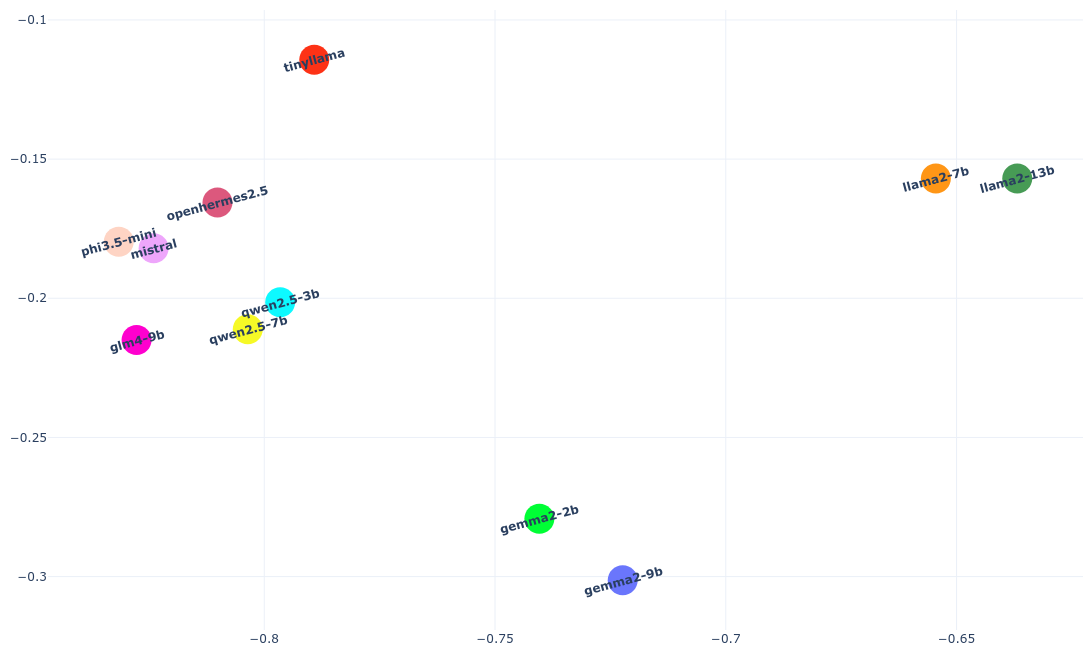}
  \caption{The visualization of PCA-2 compression of similarity matrix for eleven different LLMs.}
    \label{fig:comparison}
\end{figure}

\begin{figure}[h]
  \centering
  \includegraphics[width=0.9\linewidth]{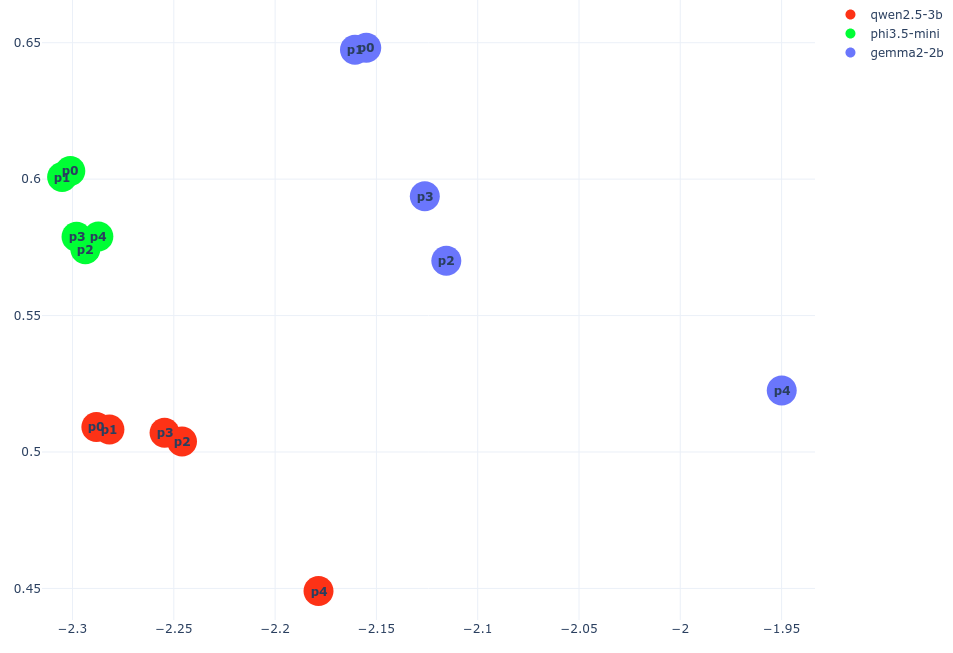}
  \caption{The visualization of PCA-2 compression of similarity matrix for three LLMs with four different system prompts and an empty system prompt.}
    \label{fig:prompt}
\end{figure}

Table \ref{tab:comparison} presents the results of comparing the previously mentioned LLMs. It shows that \textsc{ConSCompF} was able to find similarities between models trained on the same data, like the different versions of \textsc{Llama2}, \textsc{Gemma2}, and \textsc{Qwen2.5}. It also found similarities between fine-tuned versions of the same model, such as \textsc{Mistral} and \textsc{OpenHermes2.5}.

Next, we apply PCA compression to each row in the matrix, resulting in a set of two-dimensional points that represent the similarities between models (Figure \ref{fig:comparison}).

We observe that \textsc{Mistral} and \textsc{OpenHermes2.5} are close to each other on a PCA plot, since they share the same base model, while \textsc{Llama2-7b} and \textsc{Llama2-13b}, \textsc{Qwen2.5-3b} and \textsc{Qwen2.5-7b}, and \textsc{Gemma2-2b} and \textsc{Gemma2-9b} form pairs of models that share the same training data. Additionally, \textsc{TinyLlama} and \textsc{GLM4} stand out, as their training data differs from that of all other models. These results also highlight some similarity between \textsc{Mistral} and \textsc{Phi3.5-mini}, which is expected given the similarities in their \textsc{ROUGE-L} and \textsc{BLEU} metrics (Table \ref{tab:models}).

Figure \ref{fig:prompt} shows that \textsc{ConSCompF} has successfully produced a set of two-dimensional points with a clear distinction between three different models. In this PCA plot, the outputs of \textsc{Phi3.5-mini} are positioned in the top-left corner, \textsc{Qwen2.5} in the bottom-left, and \textsc{Gemma2} in the top center, with one outlier. This result suggests that, despite the variations in system prompt configurations, \textsc{ConSCompF} has successfully identified the core differences between these models.

The results indicate that \textsc{ConSCompF} has successfully identified the LLMs trained on the same data, thereby demonstrating the efficacy of the proposed framework.

\subsection{Few-shot}

\begin{table}[ht]
\centering
\begin{tabular}{lcc}
    \toprule
    Dataset&Quantization&Comparison\\
    \midrule
    Few-shot (10, cons=0.95) & 0.7682 & 0.7544 \\
    Few-shot (10, cons=0.73) & \textbf{0.9153} & 0.7156 \\
    Few-shot (10, cons=0.53) & 0.7482 & \textbf{0.8787} \\
    Few-shot (20) & 0.9661 & 0.9352 \\
    Few-shot (50) & \textbf{0.9800} & \textbf{0.9611} \\
  \bottomrule
\end{tabular}
\caption{Pearson's correlation coefficients between weighted similarity scores from processing the full dataset and processing its few-shot versions.}
\label{tab:fewshot}
\end{table}

Table \ref{tab:fewshot} demonstrates Pearson's correlation coefficients between the weighted similarity scores obtained during the experiment on the full dataset and the corresponding results obtained during few-shot runs with 10, 20, and 50 samples.

\begin{figure}[!b]
  \centering
  \includegraphics[width=0.9\linewidth]{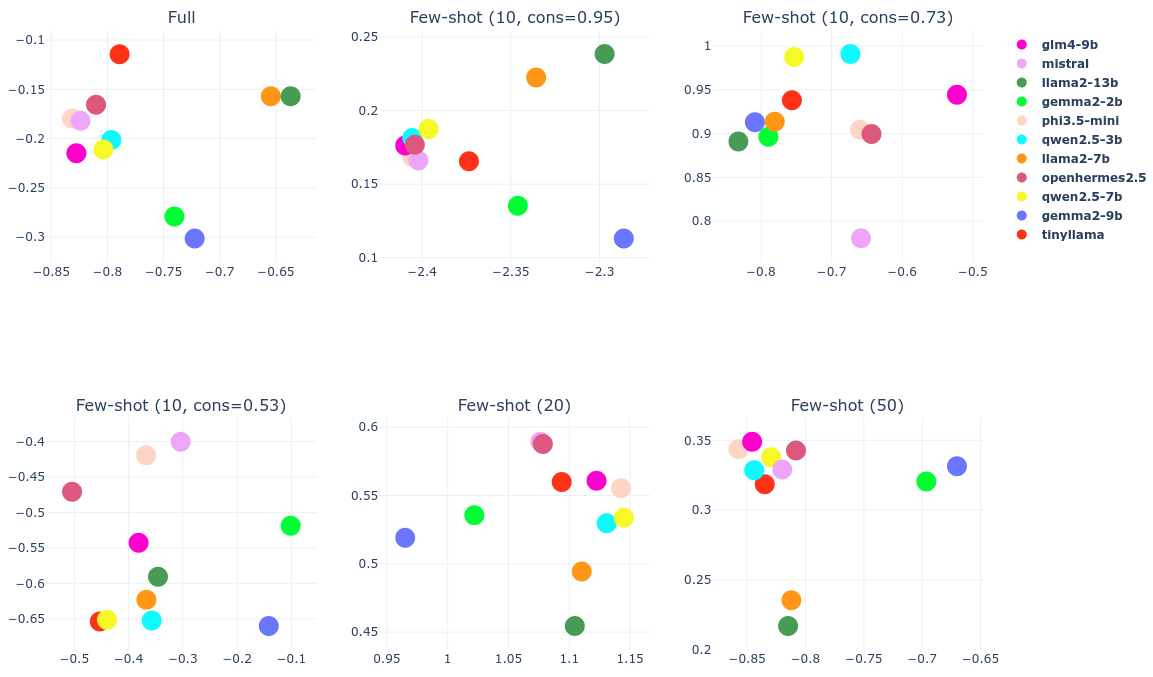}
  \caption{The visualization of PCA-2 compression of similarity matrix for eleven different LLMs in various few-shot setups.}
    \label{fig:fewshot}
\end{figure}

This data indicates that \textsc{ConSCompF} can be effectively used in a few-shot setup, as even 50 samples yield approximately the same results as 5,200 and 520 samples. Additionally, we found that the instruction consistency score has a significant impact on few-shot results. Overall, lower average instruction consistency leads to better comparison quality.

Furthermore, we apply PCA to the results of the second experiment in a few-shot setup and observe roughly the same patterns (Figure \ref{fig:fewshot}).

\subsection{Framework Efficacy}

We use Pearson's correlation coefficient (Table \ref{tab:efficacy}) to compare the weighted similarity scores to the inverted differences in ROUGE-L scores of all models in order to assess how closely the similarity scores produced by \textsc{ConSCompF} reflect the differences in the LLM performance. The results from \textsc{ConSCompF} closely align with the differences in \textsc{ROUGE-L} scores in the experiment that evaluated quantization efficacy.

\begin{table}[!t]
\centering
\begin{tabular}{lcc}
    \toprule
    Dataset&Quantization&Comparison\\
    \midrule
    Full (cons=0.85) & 0.8134 &  0.1827\\
    Few-shot (10, cons=0.95) & 0.7833 & -0.1108 \\
    Few-shot (10, cons=0.73) & \textbf{0.8834} & 0.1645 \\
    Few-shot (10, cons=0.53) & 0.5470 & \textbf{0.2522} \\
  \bottomrule
\end{tabular}
\caption{Pearson's correlation coefficient between the inverted differences in the ROUGE-L scores and \textsc{ConSCompF} similarity scores.}
\label{tab:efficacy}
\end{table}

\begin{table}[!t]
\centering
\small
\begin{tabular}{lp{2.6cm}
    *{5}{>{\centering\arraybackslash}p{1.3cm}}
    } 
    \toprule
    Dataset&Calculation\newline method&original&q8&q4&q2&pirate\\
    \midrule
    Full & Similarity & 0.8351 & 0.8384 & 0.7878 & 0.6865 & 0.6623 \\
      & \textbf{Weighted\newline Similarity} & \textbf{0.6727} &
    \textbf{0.668} & \textbf{0.5359} & \textbf{0.3533} & \textbf{0.4319} \\
    \hline
    Few-shot& Similarity & 0.7144 & 0.5460 & 0.9073 & 0.3806 & 0.5632 \\
    (10, cons=0.95)& \textbf{Weighted\newline Similarity} & \textbf{0.0815} & 
    \textbf{-0.2968} & \textbf{-0.5767} & \textbf{-0.2358} & \textbf{-0.4490} \\
    \hline
    Few-shot & Similarity & 0.6547 & 0.5845 & 0.7938 & 0.7759 & 0.5178 \\
    (10, cons=0.73)& \textbf{Weighted\newline Similarity} & \textbf{-0.1948} & \textbf{-0.1669} & \textbf{-0.0974} & \textbf{-0.1486} & \textbf{-0.3718} \\
    \hline
    Few-shot & Similarity & \textbf{-0.0986} & \textbf{0.3610} & 0.7456 & \textbf{0.5456} & \textbf{-0.0255} \\
    (10, cons=0.53)& \textbf{Weighted\newline Similarity} & -0.7363 & -0.5389 & \textbf{0.0033} & -0.5016 & -0.5998 \\
  \bottomrule
\end{tabular}
\caption{Pearson's correlation between similarity and instruction consistency in the quantization experiment.}
\label{tab:weighting}
\end{table}

Another finding suggests that the few-shot setup results, which have a lower average instruction consistency score (53\% or 73\%), are more strongly correlated with the \textsc{ROUGE-L} differences than the results from processing the full dataset (88\% for the quantization experiment and 25\% for the comparison experiment). This implies that \textsc{ConSCompF} is effective with small datasets but requires careful selection of the benchmarking instructions.

Table \ref{tab:weighting} shows the correlation between similarity scores and instruction consistency scores for LLM used in the first experiment. Overall, weighted similarity has much lower correlation with instruction consistency than its unweighted counterpart, which suggests that the weighting strategy used in \textsc{ConSCompF} is effective.

\section{Discussion}
\label{discussion}

According to the experiment results, \textsc{ConSCompF} has successfully detected the following patterns:

\begin{enumerate}
\def\labelenumi{\arabic{enumi}.}
\item
Degradation of a TinyLlama model due to quantization, which is caused by a data loss.
\item
Overall, the different versions of \textsc{TinyLlama} exhibit significant similarity, as they are all based on the same model and trained on the same data.
\item
There is a similarity between \textsc{Mistral} and \textsc{OpenHermes2.5}, as \textsc{OpenHermes2.5} is a fine-tuned version of \textsc{Mistral}. Both models share the same structure and pre-training data.
\item
Models from the same families (\textsc{Gemma2}, \textsc{Llama2}, \textsc{Qwen2.5}) exhibit high similarity because they are trained on the same data and differ only in size.
\item
On a PCA-2 plot, \textsc{TinyLlama} and \textsc{GLM4} stand out from all other models due to their unique structure, size, and training data.
\item 
Using different system prompts on the same three models yields results where all models are distinctly separated on a PCA-2 plot, despite the differences in their system prompts.
\end{enumerate}

The LLM comparison experiment reveals the patterns described above only when comparing more than two LLMs. Therefore, it is recommended to use \textsc{ConSCompF} for creating LLM similarity matrices (Table \ref{tab:comparison}) rather than for comparing individual models. Once we have a similarity matrix, we can add a new LLM to it by comparing its outputs to those of other models. This similarity matrix can also be converted into a set of points using compression algorithms such as PCA and visualized in a two-dimensional or a three-dimensional space (Figure \ref{fig:comparison}, Figure \ref{fig:fewshot}, Figure \ref{fig:prompt}).

Another observation is that similarity scores generated by \textsc{ConSCompF} in a few-shot setup with a low average instruction consistency score seem to correlate more strongly with differences in \textsc{ROUGE-L} scores than those generated from the full dataset (Table \ref{tab:fewshot}). We hypothesize that instructions with lower consistency scores require more creativity and produce less uniform responses, allowing the LLM to better demonstrate its text generation capabilities. Despite \textsc{ConSCompF}'s use of a similarity score weighting strategy to mitigate the effects of instruction consistency, instructions with high consistency scores may still yield less accurate LLM similarity scores.

To maximize the efficacy of \textsc{ConSCompF}, we recommend using instructions that prompt the model for tasks requiring a certain level of creativity. Such tasks may include writing a fiction story, devising product ideas, or writing emotional messages. Examples of instructions from a few-shot curated low-consistency set in the model comparison experiment include: "Generate a short story driven by an emotion: anger" or "Write near-future fiction set in the year 2040.". In contrast, instructions that result in high-consistency scores typically prompt the model for well-known facts or math tasks. For instance, the instruction "Name an animal that can fly." will always result in predictable answers, which are difficult to use for model comparison. Additionally, to make the comparison more efficient, it is recommended to select instructions that require complex answers with two or more sentences, ensuring the encoder has sufficient data to process.

Considering the pricing for LLM APIs, it may be difficult to create an ideal instruction set for comparing commercial LLMs. Therefore, we recommend using small, free, and open-source LLMs before applying this technique to larger commercial models with paid APIs. Finding a combination of instructions that results in the highest correlation between \textsc{ConSCompF} similarity scores and differences in scores produced by other LLM benchmarks would allow an almost zero-cost comparison of commercial LLMs.

\section{Limitations and Future Research}
\label{limitations}

Although \textsc{ConSCompF} has successfully detected the degradation of \textsc{TinyLlama} after quantization, it should not be considered an accurate LLM performance metric. \textsc{ConSCompF} only calculates similarities between models' responses, not differences in their performances. Therefore, if \textsc{ConSCompF} outputs low similarity scores for two LLMs, it does not necessarily indicate that one model outperforms the other.

In addition, as mentioned earlier, the results of the LLM comparison produced by \textsc{ConSCompF} still depend on the instruction consistency score. Therefore, the instruction list should be carefully curated before using \textsc{ConSCompF} to compare LLMs, especially in a few-shot scenario.

As shown in Table \ref{tab:encoder}, encoder selection has a great impact on the diversity of the similarity scores between LLMs. Some encoder models may not be sensitive enough for detecting the differences in LLMs' writing styles, resulting in similarity scores not being diverse enough for accurate comparison. Besides, different encoders may have different biases, emphasizing some particular aspects of input texts and ignoring the others. 

To mitigate this issue, future researchers may consider fine-tuning an underlying encoder model for improving comparison results. Encoder fine-tuning is a complex task that requires its own methodology, data processing, and experiments, all of which are out of the scope of this work. However, it has the potential to become an intriguing topic for future research. We assume that it is possible to fine-tune the encoder using the entropy of \textsc{ConSCompF} similarity scores as a training target. 

Additionally, we did not apply any specific classification algorithms for LLM categorization based on their similarity scores, which is another promising area for future research.

In conclusion, it is recommended to use \textsc{ConSCompF} as a quick metric to identify a set of the most similar LLMs, rather than as a precise benchmark for evaluating their performance.

\section{Conclusion}
\label{conclusion}

In this work, we provided an overview of the current state of text similarity comparison and LLM benchmarking and proposed \textsc{ConSCompF}, a new LLM similarity comparison framework that accounts for instruction consistency and uses adjusted weighting to reduce the correlation between consistency scores and similarity scores.

We conducted two experiments using the proposed framework and demonstrated that \textsc{ConSCompF} similarity scores correlate with differences in \textsc{ROUGE-L} scores calculated using traditional benchmarking techniques. In the first experiment, \textsc{ConSCompF} detected the post-quantization degradation of the \textsc{TinyLlama} model. In the second experiment, \textsc{ConSCompF} successfully identified similarities between LLMs trained on the same data.

Additionally, we conducted the same two experiments in several few-shot LLM comparison scenarios and found that \textsc{ConSCompF} can still produce results similar to those obtained in the experiments with a large number of prompts. However, the quality of these results depends on the consistency scores for each instruction, with lower consistency scores resulting in a higher correlation with differences in ROUGE-L scores.

Despite some of the limitations mentioned earlier, \textsc{ConSCompF} still offers many potential use cases, such as generating similarity matrices for multiple LLMs and using them for LLM categorization.

\appendix
\section*{Appendix A. Similarity Matrices}

\begin{table}[ht]
\centering
\scriptsize
\begin{tabular}{l*{11}{>{\centering\arraybackslash}p{0.65cm}}}
    \toprule
    &1&2&3&4&5&6&7&8&9&10&11 \\
    \midrule
    1. tinyllama & 1.0000 & 0.9982 & 0.9956 & 0.9978 & 0.9864 & 0.9896 & 0.9703 & 0.9935 & 0.9935 & 0.9963 & 0.9049 \\
    2. gemma2-2b & 0.9982 & 1.0000 & 0.9974 & 0.9997 & 0.9846 & 0.9877 & 0.9685 & 0.9953 & 0.9916 & 0.9945 & 0.9031 \\
    3. gemma2-9b & 0.9956 & 0.9974 & 1.0000 & 0.9978 & 0.9820 & 0.9852 & 0.9659 & 0.9979 & 0.9890 & 0.9919 & 0.9005 \\
    4. phi3.5-mini & 0.9978 & 0.9997 & 0.9978 & 1.0000 & 0.9842 & 0.9874 & 0.9682 & 0.9956 & 0.9913 & 0.9942 & 0.9028 \\
    5. glm4-9b & 0.9864 & 0.9846 & 0.9820 & 0.9842 & 1.0000 & 0.9968 & 0.9839 & 0.9799 & 0.9930 & 0.9901 & 0.9185 \\
    6. qwen2.5-3b & 0.9896 & 0.9877 & 0.9852 & 0.9874 & 0.9968 & 1.0000 & 0.9808 & 0.9830 & 0.9961 & 0.9932 & 0.9154 \\
    7. qwen2.5-7b & 0.9703 & 0.9685 & 0.9659 & 0.9682 & 0.9839 & 0.9808 & 1.0000 & 0.9638 & 0.9769 & 0.9740 & 0.9346 \\
    8. llama2-7b & 0.9935 & 0.9953 & 0.9979 & 0.9956 & 0.9799 & 0.9830 & 0.9638 & 1.0000 & 0.9869 & 0.9898 & 0.8984 \\
    9. llama2-13b & 0.9935 & 0.9916 & 0.9890 & 0.9913 & 0.9930 & 0.9961 & 0.9769 & 0.9869 & 1.0000 & 0.9971 & 0.9115 \\
    10. mistral & 0.9963 & 0.9945 & 0.9919 & 0.9942 & 0.9901 & 0.9932 & 0.9740 & 0.9898 & 0.9971 & 1.0000 & 0.9086 \\
    11. open\-hermes2.5 & 0.9049 & 0.9031 & 0.9005 & 0.9028 & 0.9185 & 0.9154 & 0.9346 & 0.8984 & 0.9115 & 0.9086 & 1.0000 \\
    \bottomrule
\end{tabular}
\caption{Inverted differences between ROUGE-L scores of LLM used in the second experiment.}
\label{tab:rougel}
\end{table}

\begin{table}[ht]
\scriptsize
\begin{tabular}{l*{11}{>{\centering\arraybackslash}p{0.65cm}}}
    \toprule
    &1&2&3&4&5&6&7&8&9&10&11 \\
    \midrule
    1. tinyllama & 1.0000 & 0.8860 & 0.8658 & 0.9065 & 0.8953 & 0.9015 & 0.8875 & 0.8853 & 0.8725 & 0.8965 & 0.9052 \\
    2. gemma2-2b & 0.8860 & 1.0000 & 0.9433 & 0.8957 & 0.9024 & 0.8969 & 0.8988 & 0.8891 & 0.8812 & 0.8876 & 0.8869 \\
    3. gemma2-9b & 0.8658 & 0.9433 & 1.0000 & 0.8801 & 0.8905 & 0.8933 & 0.8830 & 0.8786 & 0.8710 & 0.8748 & 0.8767 \\
    4. phi3.5-mini & 0.9065 & 0.8957 & 0.8801 & 1.0000 & 0.9291 & 0.9181 & 0.9177 & 0.8854 & 0.8775 & 0.9205 & 0.9175 \\
    5. glm4-9b & 0.8953 & 0.9024 & 0.8905 & 0.9291 & 1.0000 & 0.9274 & 0.9262 & 0.8811 & 0.8720 & 0.9075 & 0.9057 \\
    6. qwen2.5-3b & 0.9015 & 0.8969 & 0.8933 & 0.9181 & 0.9274 & 1.0000 & 0.9362 & 0.8970 & 0.8895 & 0.9005 & 0.9001 \\
    7. qwen2.5-7b & 0.8875 & 0.8988 & 0.8830 & 0.9177 & 0.9262 & 0.9362 & 1.0000 & 0.8911 & 0.8840 & 0.8985 & 0.9080 \\
    8. llama2-7b & 0.8853 & 0.8891 & 0.8786 & 0.8854 & 0.8811 & 0.8970 & 0.8911 & 1.0000 & 0.9419 & 0.8731 & 0.8848 \\
    9. llama2-13b & 0.8725 & 0.8812 & 0.8710 & 0.8775 & 0.8720 & 0.8895 & 0.8840 & 0.9419 & 1.0000 & 0.8680 & 0.8776 \\
    10. mistral & 0.8965 & 0.8876 & 0.8748 & 0.9205 & 0.9075 & 0.9005 & 0.8985 & 0.8731 & 0.8680 & 1.0000 & 0.9135 \\
    11. open\-hermes2.5 & 0.9052 & 0.8869 & 0.8767 & 0.9175 & 0.9057 & 0.9001 & 0.9080 & 0.8848 & 0.8776 & 0.9135 & 1.0000 \\
  \bottomrule
\end{tabular}
\caption{Weighted LLM similarity scores calculated by \textsc{ConSCompF}.}
\label{tab:comparison}
\end{table}

\vskip 0.2in
\bibliography{references}
\bibliographystyle{theapa}

\end{document}